\documentclass[10pt,twocolumn,letterpaper]{article}

\usepackage{wacv}                      

\usepackage{algorithm}     
\usepackage{algpseudocode} 
\usepackage{standalone}
\usepackage{float}
\usepackage{xcolor}
\usepackage{comment}
\usepackage{siunitx}
\usepackage{tikz}
\definecolor{wacvblue}{rgb}{0.21,0.49,0.74}
\usepackage[pagebackref,breaklinks,colorlinks,allcolors=wacvblue]{hyperref}

\def\wacvPaperID{1195} 
\def\confName{WACV}
\def\confYear{2026}

\title{FLORA: Efficient Synthetic Data Generation for Object Detection in Low-Data Regimes via finetuning Flux LoRA}

\author{
    Álvaro Patrício \quad
    Atabak Dehban \quad
    Rodrigo Ventura \\
    Institute for Systems and Robotics, University of Lisbon, Portugal \\
    {\tt\small \{alvaro.felipe, adehban, rodrigo.ventura\}@isr.tecnico.ulisboa.pt}
}

\begin{document}
\maketitle

\begin{abstract}
Recent advances in diffusion-based generative models have demonstrated significant potential in augmenting scarce datasets for object detection tasks.
Nevertheless, most recent models rely on resource-intensive full fine-tuning of large-scale diffusion models, requiring enterprise-grade GPUs~(e.g., NVIDIA V100) and thousands of synthetic images.
To address these limitations, we propose Flux LoRA Augmentation~(FLORA), a lightweight synthetic data generation pipeline.
Our approach uses the Flux 1.1 Dev diffusion model, fine-tuned exclusively through Low-Rank Adaptation~(LoRA).
This dramatically reduces computational requirements, enabling synthetic dataset generation with a consumer-grade GPU~(e.g., NVIDIA RTX 4090).
We empirically evaluate our approach on seven diverse object detection datasets.
Our results demonstrate that training object detectors with just 500 synthetic images generated by our approach yields superior detection performance compared to models trained on 5000 synthetic images from the ODGEN baseline, achieving improvements of up to 21.3\% in mAP@.50:.95. This work demonstrates that it is possible to surpass state-of-the-art performance with far greater efficiency, as FLORA achieves superior results using only 10\% of the data and a fraction of the computational cost.
This work demonstrates that a quality and efficiency-focused approach is more effective than brute-force generation, making advanced synthetic data creation more practical and accessible for real-world scenarios.
\end{abstract}
    
\section{Introduction}
\label{sec:intro}

\begin{figure}[t]
  \centering
  \includegraphics[width=\linewidth]{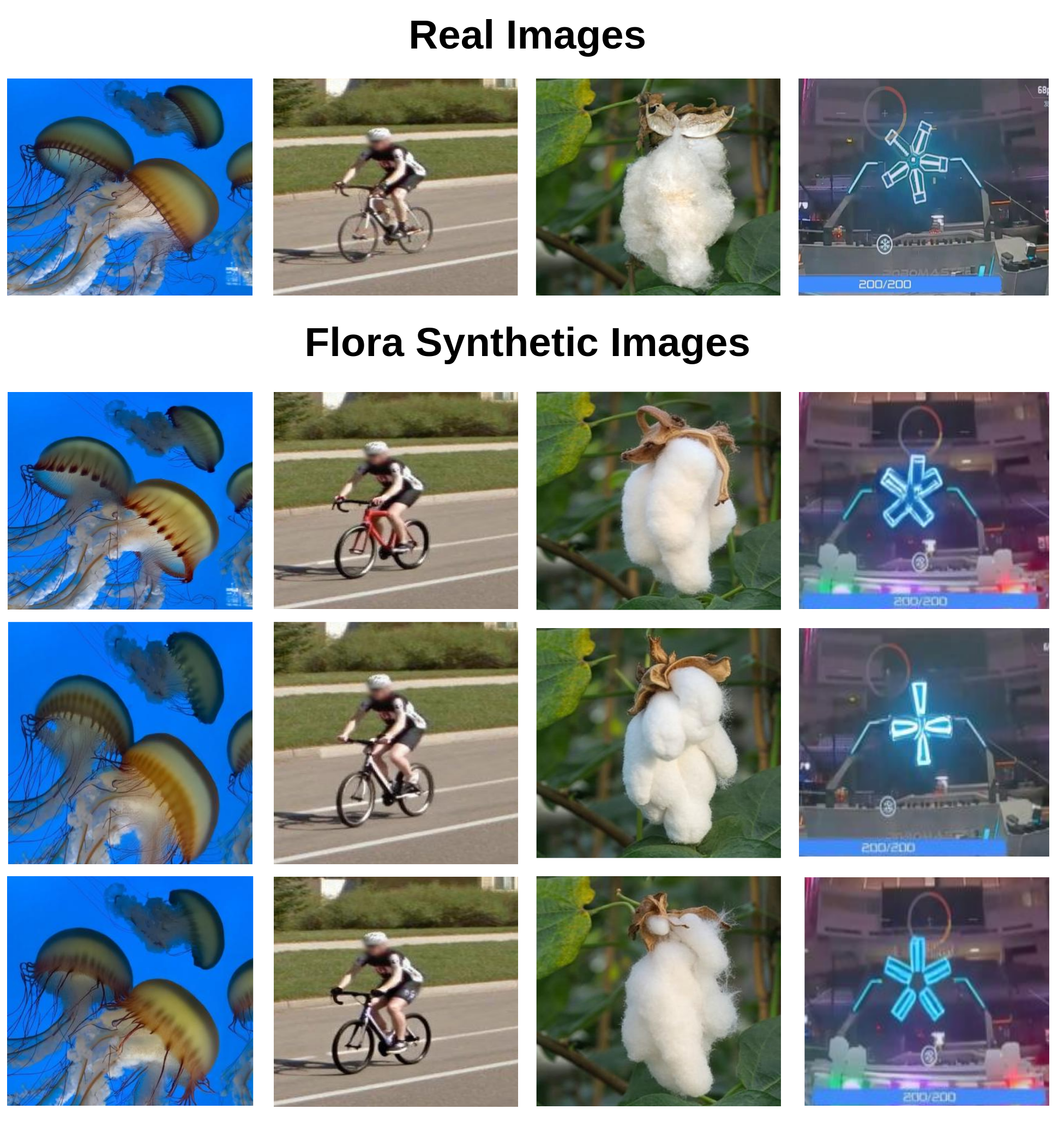} 
  \caption{Demonstration of the high-fidelity and diversity of images generated by FLORA.
For each real image~(top row), our pipeline generates multiple synthetic variations~(bottom rows).
The method's effectiveness is shown across four distinct domains—underwater, road traffic, agriculture, and gaming—from generating new poses for biological subjects~(jellyfish) and altering attributes like color~(bicycle), to creating complex organic textures~(cotton) and modifying UI elements~(game icon).}
  \label{fig:photo_collage}
\end{figure}

The performance of modern supervised deep learning models, particularly in object detection, is fundamentally constrained by the availability of large-scale, high-quality labeled datasets~\cite{deng2009imagenet}.
Obtaining such data is often expensive, time-consuming, or impractical in specialized domains like medical imaging, agricultural inspection, and environmental monitoring.
To overcome this data scarcity bottleneck, recent approaches have increasingly turned to synthetic data generation~\cite{zhu2024odgen, saharia2022imagen,zhang2023diffusionengine}.
Notably, diffusion-based models like Stable Diffusion~\cite{rombach2022stable} have emerged as exceptionally powerful tools, capable of synthesizing diverse, high-fidelity images from textual or visual prompts~\cite{gligen2023}.

However, effectively leveraging these generative capabilities for downstream object detection tasks presents its own unique set of challenges.
State-of-the-art detectors like YOLOv7~\cite{yolov7} are highly sensitive to the quality and distribution of training data~\cite{dodge2016understanding}.
For synthetic data to be beneficial, it must meet stringent criteria beyond mere photorealism.
The generated objects must be contextually plausible, geometrically precise within their bounding boxes, and stylistically consistent with the real data in terms of lighting, shadows, and texture~\cite{dehban2019impact}.
Failure to meet these requirements can introduce harmful artifacts and noise into the training process, leading a detector to learn incorrect features and ultimately degrading its performance rather than improving it.
This raises the critical need for generative pipelines that prioritize not just image quality, but also layout fidelity and contextual coherence.

This challenge is addressed by methods like ODGEN~\cite{zhu2024odgen} which leverages extensive fine-tuning of the entire Stable Diffusion model.
While powerful, this strategy requires enterprise-grade GPU resources~(e.g., NVIDIA V100) and the generation of thousands of synthetic samples, severely limiting its accessibility for researchers, smaller teams, and real-world applications with limited computational budgets.

In response to this critical gap between performance and accessibility, we propose Flux Lora Augmentation~(FLORA), a computationally lightweight and data-efficient synthetic data generation pipeline built upon the \texttt{Flux 1.1Dev} diffusion model.
We replace the costly full-model fine-tuning with a significantly more efficient Low-Rank Adaptation~(LoRA) approach~\cite{lora2021}.
This strategy drastically reduces computational demands, enabling synthetic dataset creation on a single consumer-grade GPU~(e.g., NVIDIA RTX 4090) within a day.
Furthermore, our pipeline requires an order of magnitude fewer real-world images for training compared to prior work~\cite{zhu2024odgen}, making it exceptionally suitable for low-data regimes where synthetic data is most needed.
A few examples of the diverse images produced by FLORA can be seen in Figure~\ref{fig:photo_collage}.

To rigorously validate FLORA's efficacy, we conducted extensive experiments on six diverse object detection datasets.
Our results demonstrate that a detector trained with only 500 synthetic images from our pipeline achieves superior performance compared to a model trained with 5,000 images from the ODGEN baseline, yielding mAP@.50:.95 improvements of up to 21.3\%.
As illustrated in Figure~\ref{fig:photo_collage}, our method consistently generates synthetic data with high fidelity and contextual relevance across various distinct domains.

The main contributions of this work are as follows:
\begin{itemize}
    \item We propose a lightweight synthetic data generation pipeline using a LoRA-tuned Flux 1.1Dev model, achieving state-of-the-art performance with significant improvements in computational and data efficiency.
    \item Our approach outperforms the resource-intensive ODGEN baseline on five of six benchmark datasets, using only 10\% of the synthetic data to achieve superior results.
    \item We provide the first extensive ablation study on key hyperparameters influencing LoRA-based synthetic data generation for detection, enhancing the practical applicability of our pipeline.

\end{itemize}

\section{Related Works}
\label{sec:related_works}

\paragraph{Object Detection in Low-Data Regimes.}
The performance of state-of-the-art object detectors, from ``classic architectures'' such as Faster R-CNN~\cite{ren2015faster} to modern real-time models like YOLOv7~\cite{yolov7}, is fundamentally tied to large-scale, diverse, and meticulously labeled datasets.
However, in specialized domains such as medical imaging, robotics, or agriculture, acquiring such data is often prohibitively expensive.
Traditional data augmentation techniques like MixUp~\cite{zhang2017mixup} and Copy-Paste~\cite{copypaste}, while useful, are limited to recombining existing pixel information and thus fail to introduce the true semantic and stylistic diversity needed for robust generalization.

\paragraph{Generative Models for Synthetic Data.}
To overcome these limitations, the field has increasingly turned to generative models.
While earlier paradigms like GANs~\cite{goodfellow2014gan,lan2024sustechgan} were pivotal, Denoising Diffusion Probabilistic Models (DDPMs)~\cite{ho2020ddpm} have emerged as the state-of-the-art, demonstrating unprecedented capabilities in high-fidelity image synthesis.
Foundational models such as DALL·E 2~\cite{ramesh2022dalle2}, Imagen~\cite{imagen2022}, and Stable Diffusion~\cite{stablediffusion2022} have opened a new frontier for synthetic data creation by enabling photorealistic generation from simple text prompts.

\paragraph{Diffusion Model Backbone: FLUX.1}
Our FLORA pipeline is built upon the FLUX.1 diffusion model, a modern generative backbone trained with a Flow Matching objective~\cite{flux_kontext_paper}. Distinct from traditional Denoising Diffusion Probabilistic Models (DDPMs) that learn a multi-step denoising schedule, Flow Matching learns a continuous-time probability flow that directly maps a simple noise distribution to the complex data distribution of images. This is achieved through a more direct regression objective on a vector field, simplifying the training process. Crucially, FLUX's architecture is explicitly designed for high-fidelity, in-context generation and editing tasks, making it an ideal foundation for our mask-guided inpainting pipeline, which demands strong contextual coherence.

\paragraph{Controllable Synthesis for Object Detection.}
For object detection, generic synthesis is insufficient; generation must be controllable to ensure objects appear with correct labels and in plausible locations. This need has spurred the development of powerful frameworks, from general-purpose conditioners like ControlNet~\cite{controlnet2023} and GLIGEN~\cite{gligen2023} to layout-specific models like ReCo~\cite{reco2023}, InstanceDiffusion~\cite{instancediffusion2023}, and GeoDiffusion~\cite{geodiffusion2023}. Applying these principles to the task of building complete, domain-specific datasets, ODGEN~\cite{odgen2024} has emerged as the state-of-the-art benchmark. However, while setting a new performance standard, its methodology establishes a paradigm that is not only resource-intensive but also operationally complex and less transparent. This stems from its requirement for enterprise-grade hardware, its multi-stage pipeline involving a separately trained ResNet classifier for post-generation filtering, and the lack of extensive ablation studies, which makes it challenging for researchers to reproduce or adapt the pipeline. These combined factors limit practical accessibility and leave key design choices underexplored.

\paragraph{Parameter-Efficient Fine-Tuning (PEFT).}
The immense computational cost of full fine-tuning, as exemplified by ODGEN, has catalyzed the development of Parameter-Efficient Fine-Tuning (PEFT) techniques~\cite{han2024parameter}.
Methods like LoRA~(Low-Rank Adaptation)~\cite{lora2021} have become prominent by allowing massive pretrained models to be adapted to downstream tasks by training only a tiny fraction of their parameters.
This approach is conceptually linked to personalization methods like DreamBooth~\cite{ruiz2023dreambooth} and Textual Inversion~\cite{gal2022textual}, which specialize models on new visual concepts from just a handful of images, further validating the potential of lightweight adaptation.

\paragraph{Distinction from Alternative Diffusion-Based Paradigms.}
It is crucial to distinguish our data augmentation pipeline from other works that also leverage diffusion models but for different tasks or with different core assumptions.
For instance, models like CamoDiffusion~\cite{chen2023camodiffusion} are end-to-end inference frameworks that perform segmentation by generating masks; in contrast, FLORA is a data generation tool designed to train other detectors.
A more direct comparison can be made with DiffusionEngine~\cite{zhang2023diffusionengine}, which, like FLORA, generates training data.
However, a key difference lies in the label generation process.
DiffusionEngine creates entirely new images and uses a learned ``Detection-Adapter" to produce pseudo-labels, a process that can introduce annotation noise.
FLORA's mask-guided inpainting approach, by design, surgically modifies objects within their original ground-truth bounding boxes.
This fundamental design choice guarantees the integrity of the original labels and the scene's background composition, offering a more direct and reliable method for high-fidelity data augmentation.

\paragraph{Our Contribution.}
The use of AI-generated content for data augmentation is gaining traction, with recent works successfully applying it to specialized domains like bolt detection for construction inspection~\cite{wu2025aigcbolt}. While such studies validate the promise of synthetic data, our contribution addresses the core methodological challenge of creating a \textit{general-purpose, efficient, and rigorously benchmarked} pipeline. To this end, we contrast FLORA directly with the current methodological state-of-the-art, ODGEN~\cite{odgen2024}, which relies on resource-intensive full model fine-tuning. By benchmarking against ODGEN, we systematically validate—to our knowledge, for the first time—that a lightweight, LoRA-based approach can surpass this heavy benchmark, delivering superior downstream detection performance with an order-of-magnitude less synthetic data. This finding highlights the critical role of targeted, high-fidelity generation over sheer data volume and proposes a more practical path forward for the field.
\section{Methodology}
\label{sec:3_methodology}

We introduce \textbf{FLORA} (\textit{Flux LoRA Augmentation}), a streamlined two-stage pipeline for generating high-quality synthetic data in challenging low-data scenarios.
FLORA integrates mask-guided inpainting~\cite{lugmayr2022repaint} 
with computationally efficient Low-Rank Adaptation (LoRA)~\cite{lora2021} to address key limitations in prior methods like ODGEN, enabling targeted data augmentation via large diffusion models with minimal resources.
The two stages of our pipeline are: 1) object-centric LoRA fine-tuning and 2) LoRA-conditioned synthetic image generation.

\subsection{Stage 1: Object-Centric LoRA Fine-Tuning}

\begin{figure*}[ht]
 \centering
 \includegraphics[width=0.85\textwidth]{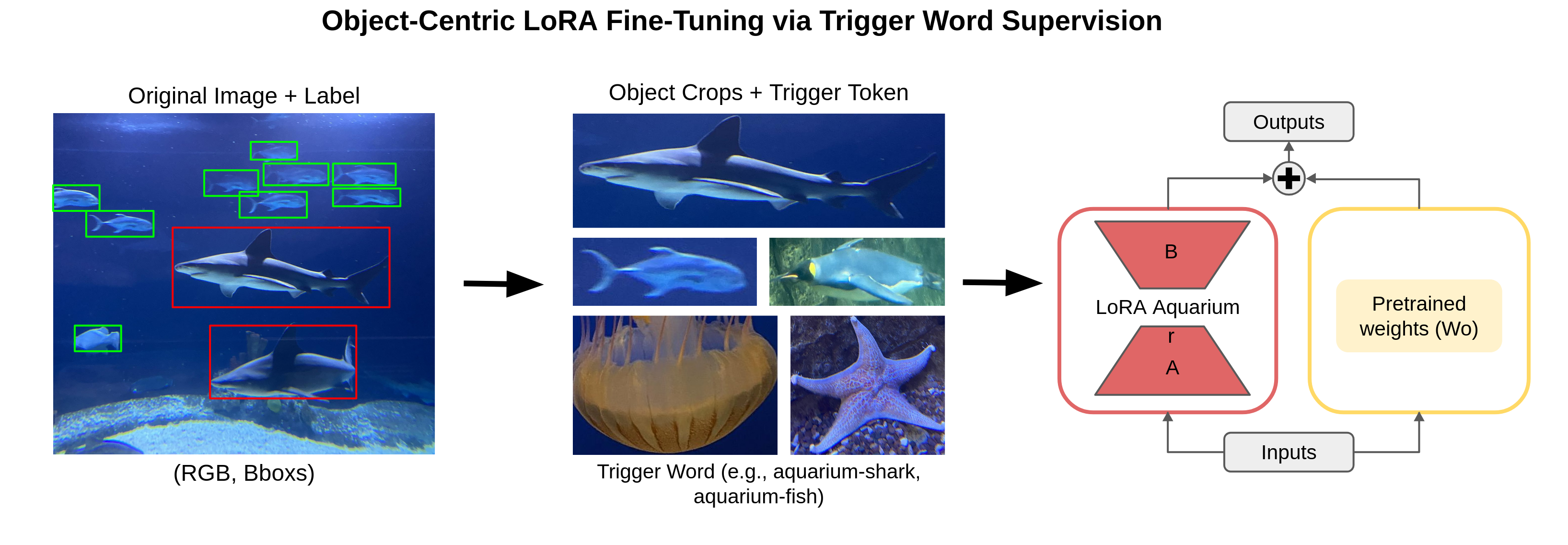}
 \caption{The FLORA fine-tuning stage. From a labeled image, object crops are extracted based on bounding box annotations. Each crop is paired with a structured trigger word (e.g., \texttt{aquarium-shark}).
 These pairs are used to fine-tune a lightweight LoRA module~(red box) injected into a frozen diffusion backbone~(yellow box), resulting in an expert generator for that specific object class.}
 \label{fig:lora_training_pipeline}
\end{figure*}

The first stage of FLORA creates a specialized, lightweight LoRA module for each object category in a dataset.
While large diffusion models like Flux excel at general domains, their ability to generate specific or technical content~(e.g., a particular species of starfish or a cancerous MRI lesion) is limited.
Our approach bypasses the need for full model retraining by efficiently adapting the pretrained model to these niche visual concepts.
The process is visualized in Figure~\ref{fig:lora_training_pipeline}.

\paragraph{Data Preparation and Trigger Words.}
For each object category, we prepare a fine-tuning dataset by extracting a small, fixed number of samples.
Specifically, we randomly select just 30 object crops per category directly from the training set's bounding box annotations.
This number was deliberately chosen to operate within a low-data regime, a paradigm validated by seminal works in generative model personalization.
Foundational methods such as DreamBooth~\cite{ruiz2023dreambooth} and Textual Inversion~\cite{gal2022textual} have demonstrated high-fidelity concept learning from as few as 3--5 images.
Our choice of 30 images therefore aligns with this principle of data efficiency while providing sufficient visual diversity to train a robust LoRA module.
This strategy enforces a strict low-data regime and focuses the training process exclusively on object-relevant pixels, as defined by:

\begin{equation}
    I_{\text{crop}}^c = I_{\text{RGB}}^c[x_{\min}:x_{\max}, y_{\min}:y_{\max}]
\end{equation}
where $I_{\text{RGB}}^c$ is a sample image containing class $c$ and the coordinates define the bounding box.
This avoids manual curation and minimizes background noise.
Each crop is then captioned with a structured trigger word, $t_w$, for precise conditioning following the format \texttt{DatasetName-ClassLabel}.
This systematic naming convention ensures reproducibility and scalability across diverse datasets.

\subsection{Stage 2: Synthetic Image Generation via LoRA-Conditioned Inpainting}
\label{sec:inference_pipeline}

\begin{figure*}[ht]
 \centering
 \includegraphics[width=0.85\textwidth]{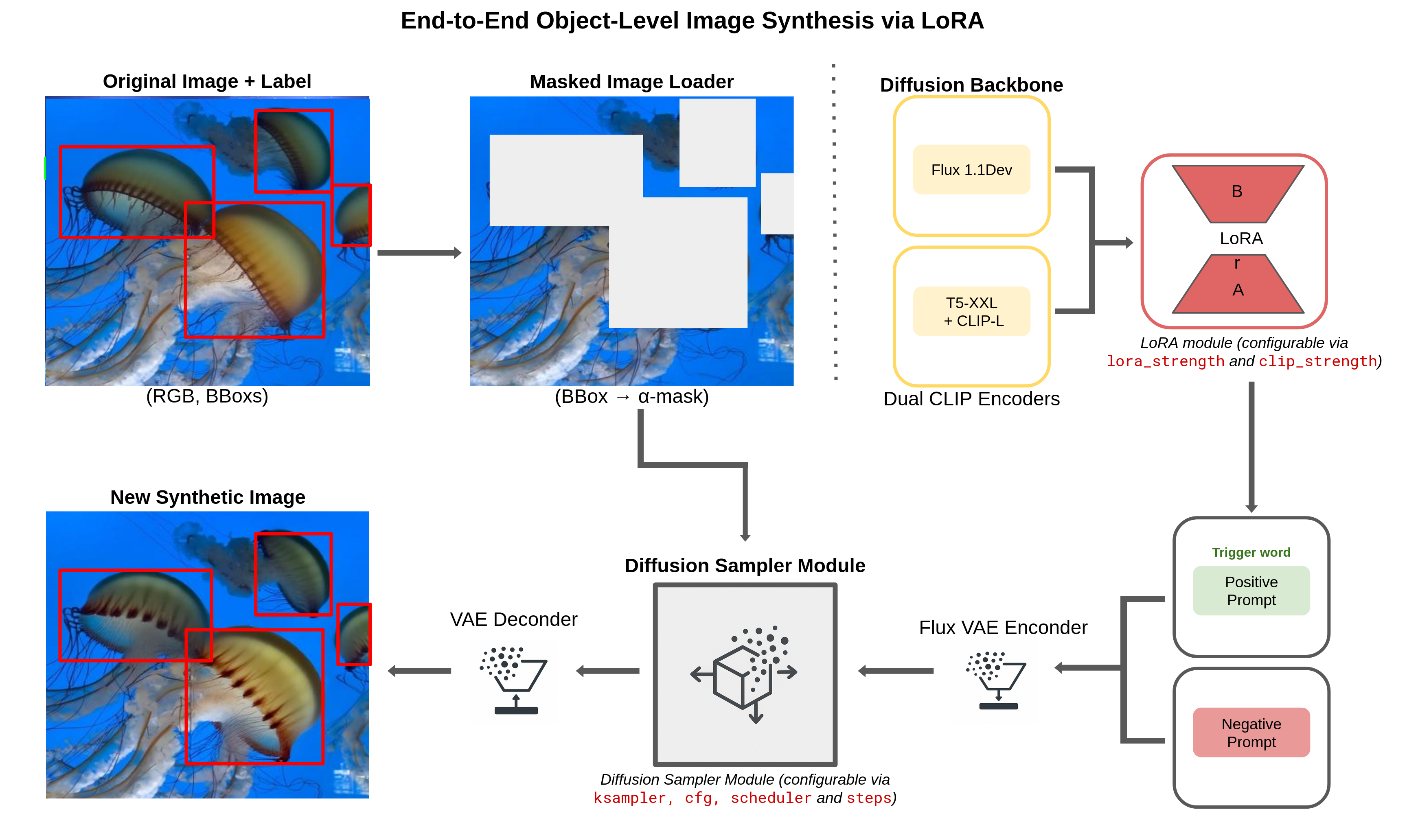}
 \caption{The FLORA generation pipeline. An input image's bounding boxes are converted into masks. The masked image and a class-specific trigger word are fed into the LoRA-adapted diffusion model.
A sampler generates new object latents within the masked regions, which a VAE decodes into a photorealistic synthetic image, preserving the original background and layout.}
 \label{fig:comfyui_pipeline}
\end{figure*}

In the second stage, the fine-tuned LoRA module from Stage 1 is used as an expert generator to produce synthetic images via a compositional inpainting process.
This approach, visually outlined in Figure~\ref{fig:comfyui_pipeline}, allows us to create new object instances that respect the original image's composition and layout, guided by a series of precise, programmatically controlled steps.

\paragraph{Input Preparation and Masking.}
The pipeline begins by taking an original training image and its corresponding YOLO annotation file as input.
Following the logic in our ``create alpha mask from yolo'' function, we parse the annotation file to identify the bounding box of the target object class.
An RGBA version of the input image is then created, where the alpha channel within this specific bounding box is set to zero, effectively creating a transparent region.
This serves as the primary inpainting mask.
To ensure a seamless blend between the original background and the newly generated object, this sharp-edged mask is subsequently processed with a Gaussian blur~(kernel size 5, sigma 5), softening the transition boundary.

\paragraph{Core Generation Pipeline.}
The masked image and a class-specific trigger word form the inputs for the core diffusion process, which proceeds as follows:
\begin{enumerate}
    \item \textbf{Text and LoRA Conditioning:} The trigger word (e.g., \texttt{underwater-starfish}) are converted into numerical embeddings by a dual text-encoder system (CLIP-L and T5-XXL). Concurrently, the pre-trained weights ($W$) of the Flux 1.1Dev are dynamically merged with the weights of the fine-tuned LoRA module ($A, B$) using the formula $W' = W + \alpha \times (A \cdot B)$, where $\alpha$ is the LoRA strength and $\cdot$ is matrix multiplication.

    \item \textbf{Latent Space Preparation:} The masked RGBA image is encoded into the latent space by the VAE encoder. The `InpaintModelConditioning' node then prepares the initial state for the sampler by combining the encoded latent image, the blurred mask, and the positive text embeddings.

    \item \textbf{Iterative Denoising (Sampling):} The actual image generation occurs in the latent space through a guided denoising process. A `karras` scheduler~\cite{karras2022elucidating} generates the noise schedule (`sigmas`), defining the steps for the `euler ancestral` sampler. Guided by the text prompts via Classifier-Free Guidance (CFG)~\cite{ho2022classifierfree}, the sampler iteratively refines the noisy latent representation exclusively within the masked region, progressively "painting in" a new object instance consistent with the trigger word.

    \item \textbf{Decoding to Pixel Space:} Once the sampling loop is complete, the final, refined latent representation is passed to the VAE decoder, which translates it back into a high-fidelity pixel space image.
\end{enumerate}

\paragraph{Workflow Orchestration and Parameters.}

This entire pipeline is executed programmatically for each source image, generating $k$ synthetic variations with different random noise seeds to ensure diversity.
Key generation hyperparameters were fixed across all experiments: a CFG scale of 6.5, and 10 sampling steps, chosen for their optimal balance of quality and speed based on our ablation studies.
Crucially, our approach simplifies the workflow by deliberately omitting any post-generation filtering as opposed to the ODGEN pipeline, which requires an additional step of training and deploying a ResNet-based image classifier~\cite{he2016deep} to validate the quality of its generated bounding boxes.
In our method, the generated image is saved directly, and the original YOLO label file is copied to accompany it~(because the label and the bounding box do not change).
This entire workflow, from mask creation to final image saving, is orchestrated by a custom Python script, which interfaces with the robust node-based backend of ComfyUI~\cite{comfyui} to manage the complex data flow.
\section{Experiments}
\label{sec:4_experiments}

\begin{figure}[ht]
  \centering
  \includegraphics[width=\linewidth]{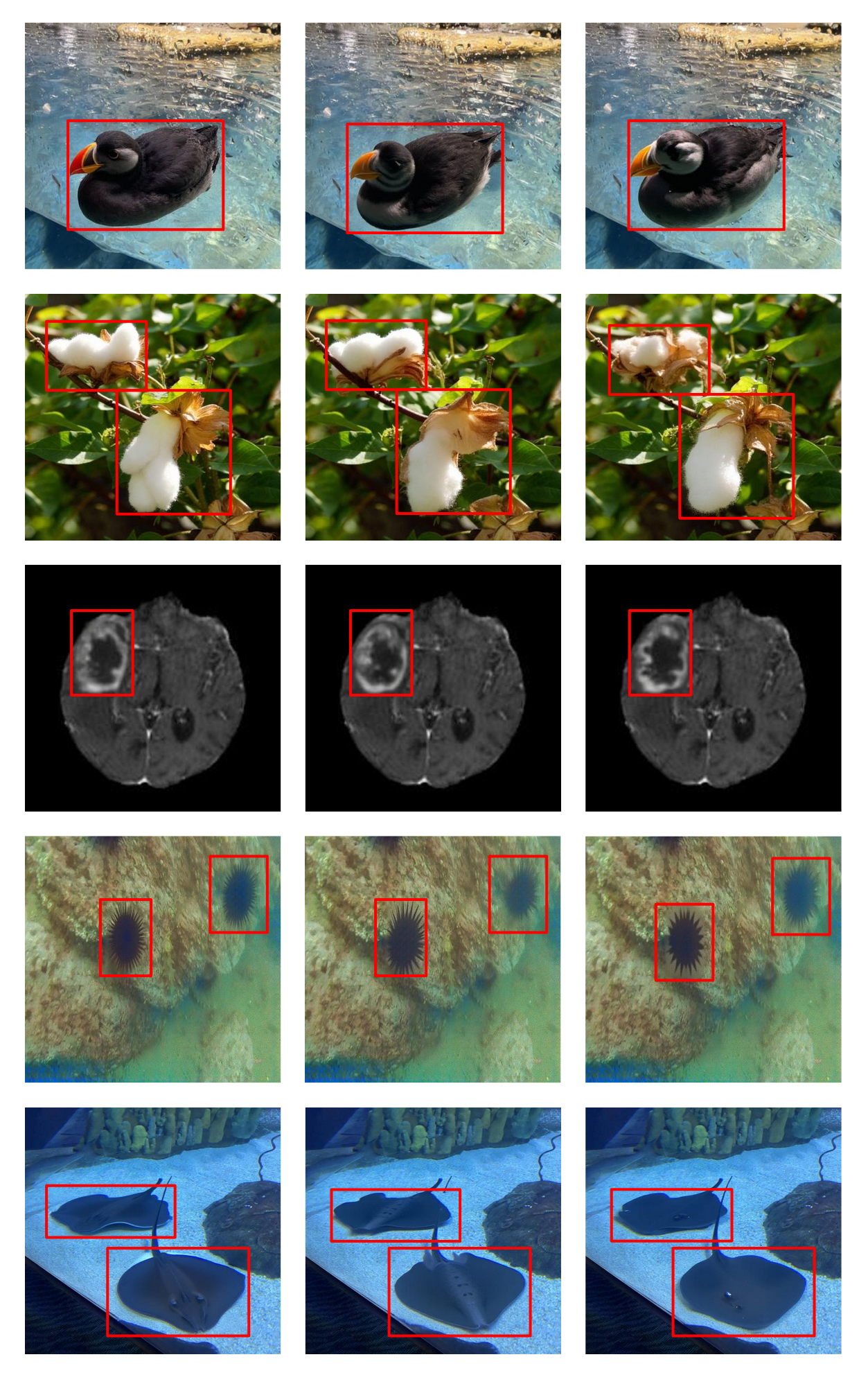}
  \caption{A showcase of synthetic images generated by the FLORA pipeline across five diverse domains: puffin, cotton, MRI brain scans, sea urchins, and stingrays. Each row presents three unique synthetic variations for a given scene, demonstrating FLORA's ability to produce high-fidelity and varied data suitable for object detection training, as indicated by the bounding boxes.}
  \label{fig:flora_results}
\end{figure}

We conducted a comprehensive set of experiments to validate the effectiveness and efficiency of FLORA.
Our evaluation is designed to answer two primary questions: (1) Does our lightweight synthetic data improve downstream object detection performance compared to state-of-the-art methods? and (2) How does the perceptual quality of our generated images, showcased in Figure~\ref{fig:flora_results}, compare to more computationally intensive approaches?

\subsection{Experimental Setup}

\paragraph{Datasets and Protocol.}
Our evaluation uses six diverse, domain-specific datasets from the Roboflow 100 benchmark: \textit{RoboMaster}, \textit{MRI}, \textit{Cotton}, \textit{Road Traffic}, \textit{Aquarium}, and \textit{Underwater} \cite{roboflow100}.
To ensure a fair and reproducible comparison, we strictly adhere to the experimental setup established in the ODGEN paper~\cite{zhu2024odgen}, adopting their official training protocol and using the same dataset configurations, for which we refer the reader to their supplemental material for a full breakdown.
All object detection models were trained for 100 epochs using a YOLOv7 detector.
We focused exclusively on YOLOv7 due to our computational budget and the fact that, as demonstrated in the ODGEN study, its performance trends are nearly identical to YOLOv5, consistently yielding slightly higher mAP scores.

\paragraph{Evaluation Metrics.}
We assess performance using two standard metrics.
For downstream object detection, we report the mean Average Precision (mAP) over IoU thresholds from 0.50 to 0.95 (mAP@[.50:.95]).
For image quality and realism, we employ the Fréchet Inception Distance (FID)~\cite{heusel2017gans}, where lower scores indicate a closer perceptual match between the real and synthetic image distributions.

\paragraph{Baselines and SOTA Comparison.}
We benchmark FLORA against a no-augmentation baseline~(training only on real images from the dataset) and six recent synthetic data generation methods: ReCo, GLIGEN, ControlNet, GeoDiffusion, InstanceDiffusion, and ODGEN.
To maintain the integrity of our comparison, we first reproduced the baseline mAP results reported in the ODGEN paper for each dataset.
We excluded the \textit{Apex Game} dataset from our study because we were unable to reproduce its reported baseline; our best effort yielded an mAP of 36\%, a significant deviation from the 47\% reported by ODGEN.
By focusing on the six datasets where the baseline was successfully verified, we ensure our comparative analysis is robust and fair.

\paragraph{Computational Efficiency as a Core Constraint.}
To directly challenge the paradigm that synthetic data generation requires extensive resources, FLORA's methodology was constrained to a highly efficient budget. We limited the synthetic set to 500 images per dataset, a 10-fold reduction from the 5,000 used by ODGEN, and ensured the entire generation process could be executed on a single consumer-grade GPU. All performance gains reported in this paper were achieved under these practical constraints, demonstrating the viability of our lightweight approach.

\paragraph{Implementation Details.}
The LoRA fine-tuning targets the attention layers of a Flux-based diffusion model (\texttt{flux1-dev.safetensors}).
We used a LoRA configuration with a rank~(dimension) of 32 and an alpha of 16.
Textual conditioning was managed by dual CLIP-L~\cite{radford2021clip} 
and T5-XXL~\cite{raffel2020t5} 
encoders.
Training was performed on $512 \times 512$ pixel images for 5 epochs using an 8-bit AdamW optimizer~\cite{dettmers2022llm} 
with \texttt{bfloat16} mixed precision.
On a consumer-grade NVIDIA RTX 4090 GPU, this process takes approximately 9--12 hours per LoRA module, demonstrating the method's high efficiency.

\subsection{Quantitative Results and Analysis}

We now present the main results, comparing FLORA against existing methods on both image fidelity and downstream detection performance.

\paragraph{Image Fidelity Analysis (FID).}
As shown in Table~\ref{tab:fid_results}, FLORA achieves state-of-the-art or highly competitive FID scores across all six benchmarks, despite using only 500 generated samples for the calculation.
Notably, it secures the best (lowest) FID on five out of the six datasets, with dramatic improvements in challenging domains like \textit{MRI} (27.60 vs. 93.82 from ODGEN) and \textit{Aquarium} (56.18 vs. 83.07).
This stability, even with a smaller sample size, is mathematically justifiable.
The Fréchet Inception Distance is defined as:
\begin{equation}
\label{eq:fid_definition}
\mathrm{FID}(x, g) = \|\mu_x - \mu_g\|_2^2 + \mathrm{Tr}(\Sigma_x + \Sigma_g - 2(\Sigma_x \Sigma_g)^{1/2})
\end{equation}
where $(\mu, \Sigma)$ are the mean and covariance of the Inception feature distributions of real ($x$) and generated ($g$) images.
According to analyses of the metric's statistical properties~\cite{chong2020pitfalls}, the sampling variance of its empirical estimate, $\widehat{\mathrm{FID}}$, can be approximated to the first order as:
\begin{equation}
\label{eq:fid_variance}
\mathrm{Var}[\widehat{\mathrm{FID}}] \approx \frac{A}{n_r} + \frac{B}{n_g}
\end{equation}
where $n_r$ and $n_g$ are the number of real and generated images, and $A, B$ are dataset-specific constants.
In our setup, $n_r = 200$ is fixed.
When comparing our approach ($n_g = 500$) with the standard benchmark ($n_g = 5,000$), the change in variance is governed solely by the second term.
As this term is often not the dominant source of variance in low-data scenarios, our empirical tests confirm that the 95\% confidence interval only tightens marginally (from $\pm 0.24$ to $\pm 0.20$).
This negligible gain in metric precision does not justify the 10x increase in computational cost, validating our lightweight approach.

\begin{table*}[t]
  \centering
  \caption{FID (↓) scores computed for images synthesized by each approach.
Lower values indicate better perceptual quality.
Our method, FLORA, was evaluated using only 500 synthetic images, whereas all other methods used 5000.
Despite this, FLORA achieves superior results on 4 of the 6 datasets, highlighting its high-fidelity generation.}
  \label{tab:fid_results}
  \begin{tabular}{l|cccccc|c}
    \toprule
    \textbf{Dataset} & ReCo & GLIGEN & ControlNet & GeoDiffusion & ODGEN & \textbf{FLORA} \\
    \midrule
    Robomaster   & 70.12  & 167.44 & 134.92 & 76.81  & \textbf{57.32} &  61.54\\
    MRI          & 202.36 & 270.52 & 212.45 & 341.74  & 93.82 & \textbf{27.60} \\
    Cotton       & 108.55 & 89.85  & 196.87 & 203.02  & 85.17 & \textbf{80.48} \\
    Road Traffic & 80.18  & 98.83  & 162.27 & 68.11    & 63.52 & \textbf{55.54} \\
    Aquarium     & 122.71 & 98.38  & 146.26 & 162.19   & 83.07 & \textbf{56.18} \\
    Underwater   & 73.29  & 147.33 & 126.58 & 125.32 &   \textbf{70.20} & 70.35 \\
    \bottomrule
  \end{tabular}
\end{table*}

\paragraph{Object Detection Performance (mAP).}
The primary results of our study are presented in Table~\ref{tab:training_results}.
Augmenting the real training images with just 500 synthetic samples from FLORA leads to substantial performance gains, outperforming the baseline and all competing state-of-the-art methods on five of the six datasets.
The performance lift is particularly striking for \textit{Road Traffic}, where FLORA achieves an mAP of 62.3\%, an increase of 21.3 points over the baseline and 18.5 points over the next-best method, ODGEN.
On the \textit{MRI} dataset, while ODGEN achieves a slightly higher mAP, FLORA remains highly competitive and significantly surpasses all other approaches.
These results demonstrate that a large quantity of synthetic data is not a prerequisite for high performance; rather, a smaller, targeted, and higher-quality set of examples can be more effective, validating FLORA's core design philosophy.

\begin{table*}[t]
  \centering
  \caption{mAP@[.50:.95] (↑) of YOLOv7 on the six benchmark datasets.
All SOTA models are trained with 200 real + 5000 synthetic images.
Our method, FLORA, uses only 200 real + 500 synthetic images, yet outperforms all competitors on 5 of the 6 datasets.}
  \label{tab:training_results}
  \begin{tabular}{l|c|cccccc}
    \toprule
    \textbf{Dataset} & \textbf{Baseline} & \textbf{ReCo} & \textbf{GLIGEN} & \textbf{ControlNet} & \textbf{GeoDiffusion}  & \textbf{ODGEN} & \textbf{FLORA} \\
    \midrule
        \textit{real + synth} & $200+0$ & $200+5\,000$ & $200+5\,000$ & $200+5\,000$ & $200+5\,000$ & $200+5\,000$ & $200+500$  \\
    \midrule
    Robomaster   & 26.5 & 27.9 & 25.0 & 32.9 & 22.6  & 34.7 & \textbf{38.2} \\
    Cotton       & 20.5 & 37.5 & 39.0 & 35.1 & 36.0  & 43.2 & \textbf{44.5} \\
    Road Traffic & 41.0 & 29.3 & 29.5 & 30.5 & 29.4  & 43.8 & \textbf{62.3} \\
    Aquarium     & 29.6 & 34.3 & 32.2 & 25.6 & 30.9  & 38.5 & \textbf{41.0} \\
    MRI          & 27.4 & 38.3 & 25.9 & 37.2 & 38.9  & \textbf{41.5} & 37.6 \\
    Underwater   & 19.4 & 15.8 & 18.5 & 17.8 & 17.2 & 22.0 & \textbf{24.5} \\
    \bottomrule
  \end{tabular}
\end{table*}

\subsection{Ablation Studies}
\label{sec:ablation_study}

To validate our hyperparameter choices and understand their impact, we conducted a series of ablation studies.
We focused our analysis on the two most challenging datasets, which exhibited the lowest baseline mAP and highest FID scores: Underwater (due to severe color shifts and low contrast) and MRI (as it lies far outside the natural image domain of the base diffusion model).

\begin{figure}[ht]
  \centering
  \includegraphics[width=\linewidth]{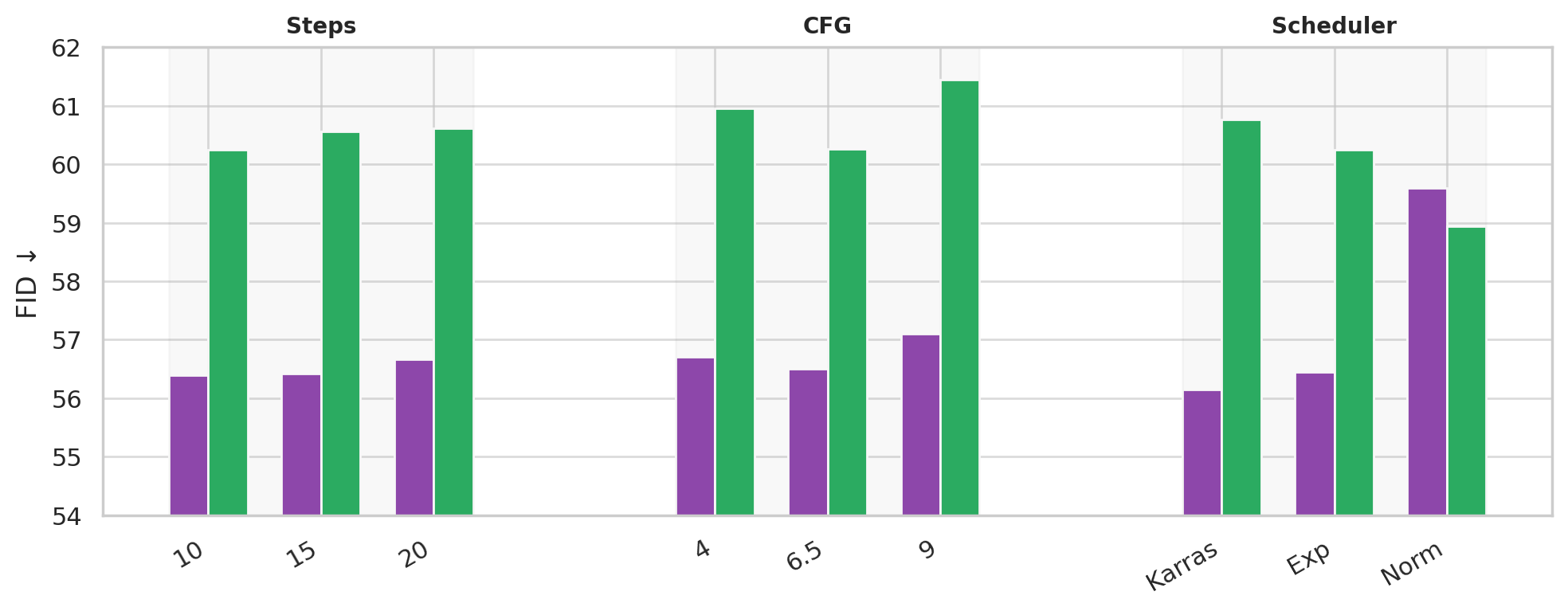}
  \caption{FID (↓) ablation on the \textit{Underwater} dataset.
Bars show fidelity for the \textit{Echinus} (purple) and \textit{Starfish} (green) classes while varying (a) sampling steps, (b) CFG scale, and (c) latent–noise scheduler.
Lower values indicate better image quality; the Karras scheduler, 10-step, and CFG=6.5 setting achieve the best trade-off.}
  \label{fig:ablation_underwater_b}
\end{figure}

\begin{figure}[ht]
  \centering
  \includegraphics[width=\linewidth]{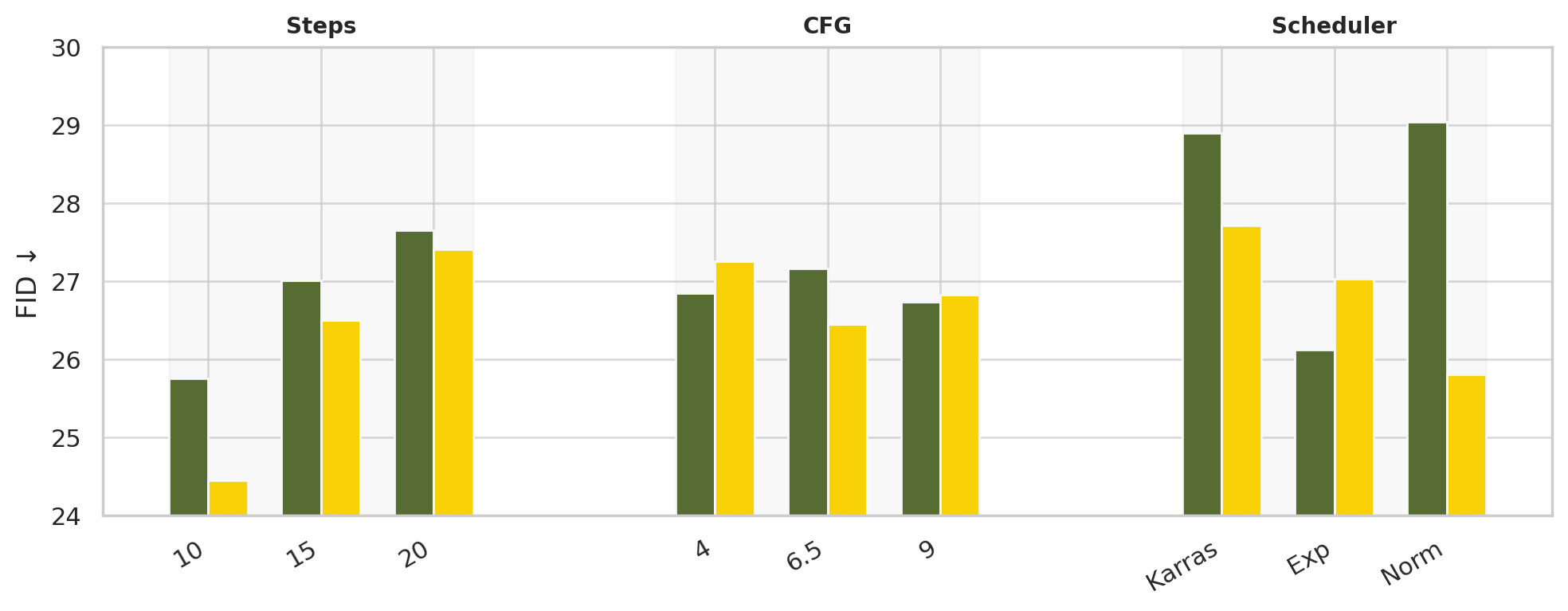}
  \caption{FID (↓) ablation on the \textit{MRI} dataset.
FID is reported for \textit{Negative} (green) and \textit{Positive} (yellow) tumor slices under the same hyperparameter sweep.
Results confirm that 10 sampling steps, CFG=6.5, and a Karras scheduler yield the most consistent fidelity.}
  \label{fig:ablation_mri_b}
\end{figure}

\paragraph{Global Generation Hyperparameters.}
As shown in Figures~\ref{fig:ablation_underwater_b} and~\ref{fig:ablation_mri_b}, we measured FID while varying key generation parameters. We swept over (i) the number of sampling steps, (ii) the classifier-free guidance (CFG) scale, and (iii) the latent-noise scheduler~\cite{zhang2023fast}. The trends remained largely consistent across both datasets. We found that FID generally improved with more sampling steps but with diminishing returns, and selected 10 steps as an optimal trade-off between image quality and generation speed. A CFG scale of 6.5 consistently provided the best balance between prompt fidelity and image realism. For the scheduler, both the Karras and Exponential options produced highly competitive results, with their performance on the MRI dataset being nearly indistinguishable. However, given its faster generation time, the Karras scheduler offered the best overall balance of quality and efficiency. Based on this analysis, we selected the combination of 10 steps, CFG=6.5, and the Karras scheduler as our default configuration

\paragraph{LoRA-Specific Factors and Prompt Complexity.}
The investigation of LoRA-specific parameters, presented in Table~\ref{tab:ablation_lora_prompt}, reveals that our pipeline is robust to minor variations.
We varied the LoRA strength and clip values and found they produced only second-order changes, with FID varying by less than 1.0 within each class.
Furthermore, we compared our minimalist prompt (the trigger word alone) with a “complex” prompt that included a descriptive sentence (e.g., “a low-resolution underwater photograph of a brittle starfish on the seabed”).
The complex prompt offered no systematic improvement.
This confirms that our trigger-word-only approach is sufficient for high-quality synthesis and that our pipeline is not overly sensitive to LoRA-specific hyperparameters.

\begin{table}[t]
\centering
\small
\setlength{\tabcolsep}{4pt}
\caption{FID (↓) for the two underwater classes under different LoRA and prompt configurations.
Results show minimal sensitivity to LoRA strength, LoRA clip, and prompt complexity, validating our simplified approach.}
\label{tab:ablation_lora_prompt}
\begin{tabular}{p{3cm} c c}
\toprule
\textbf{Configuration} & \textbf{Echinus} & \textbf{Starfish} \\
\midrule
\multicolumn{3}{@{}l}{\textbf{LoRA Strength}} \\
\hspace{1em}0.75 & 64.706 & 74.186 \\
\hspace{1em}1.00 (Default) & 65.177 & 74.071 \\
\hspace{1em}1.25 & 65.141 & 73.508 \\
\hspace{1em}1.50 & 65.945 & 74.583 \\
\midrule
\multicolumn{3}{@{}l}{\textbf{LoRA Clip}} \\
\hspace{1em}0 & 65.011 & 73.601 \\
\hspace{1em}0.50 & 65.719 & 74.133 \\
\hspace{1em}1.00 (Default) & 65.169 & 74.022 \\
\hspace{1em}1.25 & 64.991 & 73.865 \\
\hspace{1em}1.50 & 66.155 & 74.234 \\
\midrule
\multicolumn{3}{@{}l}{\textbf{Prompt Complexity}} \\
\hspace{1em}Complex & 64.635 & 74.136 \\
\hspace{1em}Normal (Default) & 65.171 & 73.771 \\
\bottomrule
\end{tabular}
\end{table}

\paragraph{Takeaways from Ablations.}
Our ablation studies demonstrate that: (i) a modest budget of 10 sampling steps with a CFG of 6.5 and a Karras scheduler provides robust performance across disparate domains; (ii) FLORA is not sensitive to minor variations in LoRA-specific parameters or prompt verbosity, simplifying its application; and (iii) the consistent FID gap between classes (e.g., Echinus vs.\ Starfish) aligns with their respective mAP scores, confirming FID’s proxy effectiveness.

\section{Conclusion and Future Work}
\label{sec:5_conclusion}

In this work, we introduced FLORA, a lightweight and efficient pipeline designed to address the critical trade-off between performance and accessibility in synthetic data generation. By coupling a modern Flux 1.1Dev diffusion backbone with parameter-efficient Low-Rank Adaptation (LoRA), we demonstrated that high-quality, domain-specific data augmentation is achievable on consumer-grade hardware. Our extensive experiments across six challenging benchmarks showed that training a YOLOv7 detector with just 500 synthetic images generated by FLORA consistently outperforms models trained with 5,000 images from the resource-intensive ODGEN baseline. This resulted in significant downstream performance gains of up to 21.3\% in mAP@.50:.95 while simultaneously setting a new state-of-the-art in perceptual quality as measured by FID scores.

Our findings present compelling evidence that challenges the prevailing "more is better" paradigm in synthetic data generation, proving that a smaller, higher-quality set of examples is more effective for training robust object detectors. The success of FLORA is attributable to its efficient object-centric fine-tuning and precise mask-guided inpainting, which ensure both semantic and geometric fidelity. While our comprehensive ablation studies provide a practical roadmap for immediate implementation, looking ahead, we believe further gains are achievable. The object-centric LoRA fine-tuning stage was intentionally kept minimal (30 crops, 5 epochs) to emphasize accessibility; a more systematic study of LoRA training regimes (e.g., rank, learning rate, crop count) represents a promising direction to enhance quality while preserving the core efficiency of our method.

{
    \small
    \bibliographystyle{ieeenat_fullname}
    \bibliography{main}
}

\end{document}